\pdfoutput=1

\documentclass[11pt]{article}

\usepackage{emnlp2021}

\usepackage{times}
\usepackage{latexsym}
\usepackage{booktabs}
\usepackage{multirow}
\usepackage{todonotes}
\usepackage{footnote}
\makeatletter
\newcommand*\iftodonotes{\if@todonotes@disabled\expandafter\@secondoftwo\else\expandafter\@firstoftwo\fi}  
\makeatother

\usepackage[T1]{fontenc}

\usepackage[utf8]{inputenc}

\usepackage{microtype}
\usepackage{xspace}
\usepackage{graphicx}

\usepackage{amssymb}
\usepackage{amsmath}

\makeatletter
\def\blfootnote{\xdef\@thefnmark{}\@footnotetext}
\makeatother

\newcommand{\specific}{$\mathcal{S}$\xspace}

\newcommand{\multilingual}{\texttt{multi}\xspace}
\newcommand{\monolingual}{\texttt{mono}\xspace}
\newcommand{\monolingualupper}{\texttt{Mono}\xspace}

\newcommand{\generalproperty}{\texttt{general}\xspace}
\newcommand{\specificproperty}{\texttt{specific}\xspace}
\newcommand{\base}{\textit{base}\xspace}
\newcommand{\basembert}{\textsc{mBERT}\xspace}

\newcommand{\dapting}{domain adaptive pretraining\xspace}
\newcommand{\mdapting}{multilingual domain adaptive pretraining\xspace}
\newcommand{\MDAPT}{\textsc{mDAPT}\xspace}
\newcommand{\dapted}{domain adaptive pretrained\xspace}

\newcommand{\adxen}{\textsc{E$_{\text{D}}$}\xspace}
\newcommand{\adxmd}{\textsc{M$_{\text{D}}$+E$_{\text{D}}$}\xspace}
\newcommand{\adxmwiki}{\textsc{M$_{\text{D}}$+M$_{\text{WIKI}}$}\xspace}
\newcommand{\adxmwikishort}{\textsc{M$_{\text{D}}$+M$_{\text{W}}$}\xspace}
\newcommand{\multispecificmodel}{\textsc{mDAPT}\xspace}

\newcommand{\bert}{\textsc{BERT}\xspace}
\newcommand{\roberta}{\textsc{RoBERTa}\xspace}
\newcommand{\berten}{\textsc{en-BERT}\xspace}
\newcommand{\bertfr}{\textsc{fr-BERT}\xspace}
\newcommand{\bertro}{\textsc{ro-BERT}\xspace}
\newcommand{\bertes}{\textsc{es-BERT}\xspace}
\newcommand{\bertde}{\textsc{de-BERT}\xspace}
\newcommand{\bertda}{\textsc{da-BERT}\xspace}
\newcommand{\bertpt}{\textsc{pt-BERT}\xspace}
\newcommand{\biobert}{\textsc{en-bio-bert}\xspace}
\newcommand{\finbert}{\textsc{en-fin-bert}\xspace}
\newcommand{\ptbiobert}{\textsc{pt-bio-bert}\xspace}

\newcommand{\quaero}{\textsc{quaero}\xspace}

\newcommand{\bioro}{\textsc{bioro}\xspace}
\newcommand{\ncbi}{\textsc{ncbi disease}\xspace}
\newcommand{\ncbishort}{\textsc{ncbi}\xspace}
\newcommand{\pharmaconer}{\textsc{pharmaconer}\xspace}
\newcommand{\pharmaconershort}{\textsc{phar}\xspace}
\newcommand{\ptner}{\textsc{clinpt}\xspace}
\newcommand{\ptnershort}{\textsc{clin}\xspace}

\newcommand{\FPB}{\textsc{Financial PhraseBank}\xspace}
\newcommand{\FPBshort}{\textsc{PhraseBank}\xspace}
\newcommand{\DanFin}{\textsc{FinNews}\xspace}
\newcommand{\GMPC}{\textsc{One Million Posts}\xspace}
\newcommand{\GMPCexp}{\textsc{OMP}\xspace}
\newcommand{\PwCCPT}{\textsc{FinMultiCorpus}\xspace}

\newcommand{\FPBtable}{\textsc{phr.bank}\xspace}
\newcommand{\DanFintable}{\textsc{finnews}\xspace}
\newcommand{\omptwotable}{\textsc{omp-2}\xspace}
\newcommand{\ompninetable}{\textsc{omp-9}\xspace}

\newcommand{\zh}{\emph{zh}\xspace}
\newcommand{\da}{\emph{da}\xspace}
\newcommand{\nl}{\emph{nl}\xspace}
\newcommand{\fr}{\emph{fr}\xspace}
\newcommand{\de}{\emph{de}\xspace}
\newcommand{\ita}{\emph{it}\xspace}
\newcommand{\ja}{\emph{ja}\xspace}
\newcommand{\no}{\emph{no}\xspace}
\newcommand{\pt}{\emph{pt}\xspace}
\newcommand{\ru}{\emph{ru}\xspace}
\newcommand{\es}{\emph{es}\xspace}
\newcommand{\sv}{\emph{sv}\xspace}
\newcommand{\en}{\emph{en}\xspace}
\newcommand{\tr}{\emph{tr}\xspace}
\newcommand{\ro}{\emph{ro}\xspace}

\title{\textsc{mDAPT}: Multilingual Domain Adaptive Pretraining in a Single Model}

\author{Rasmus K{\ae}r J{\o}rgensen$^{1,2}$ \and Mareike Hartmann$^{3*}$ \and Xiang Dai$^{1}$ \and Desmond Elliott$^{1}$ \\
$^1$University of Copenhagen, Denmark \\
$^2$PricewaterhouseCoopers (PwC), Denmark \\
$^3$German Research Center for Artificial Intelligence (DFKI), Germany \\
\texttt{rasmuskj,xiang.dai,de@di.ku.dk} \\
\texttt{mareike.hartmann@dfki.de}
}

\begin{document}
\maketitle
\begin{abstract}

Domain adaptive pretraining, i.e. the continued unsupervised pretraining of a language model on domain-specific text, improves the modelling of text for downstream tasks within the domain. 
Numerous real-world applications are based on domain-specific text, e.g. working with financial or biomedical documents, and these applications often need to support multiple languages. However, large-scale domain-specific multilingual pretraining data for such scenarios can be difficult to obtain, due to regulations, legislation, or simply a lack of language- and domain-specific text. One solution is to train a single multilingual model, taking advantage of the data available in as many languages as possible.
In this work, we explore the benefits of domain adaptive pretraining with a focus on adapting to multiple languages within a specific domain. We propose different techniques to compose pretraining corpora that enable a language model to both become domain-specific and multilingual.
Evaluation on nine domain-specific datasets---for biomedical named entity recognition and financial sentence classification---covering seven different languages show  that a single multilingual domain-specific model can outperform the general multilingual model, and performs close to its monolingual counterpart. This finding holds across two different pretraining methods, adapter-based pretraining and full model pretraining.

\end{abstract}

\section{Introduction}

\blfootnote{$^*$The research was carried out while the author was employed at the University of Copenhagen.} The unsupervised pretraining of language models on unlabelled text has proven useful to many natural language processing tasks. The success of this approach is a combination of deep neural networks \cite{NIPS2017_3f5ee243}, the masked language modeling objective \cite{devlin-etal-2019-bert}, and large-scale corpora \cite{zhu2015aligning}. In fact, unlabelled data is so important that better downstream task performance can be realized by pretraining models on more unique tokens, without repeating any examples, instead of iterating over smaller datasets \cite{JMLR:v21:20-074}. When it is not possible to find vast amounts of unlabelled text, a better option is to continue pretraining a model on domain-specific unlabelled text \cite{han-eisenstein-2019-unsupervised,dai-etal-2020-cost}, referred to as \dapting \cite{gururangan-etal-2020-dont}. This results in a better initialization for consequent fine-tuning for a downstream task in the specific domain, either on target domain data directly \cite{gururangan-etal-2020-dont}, or if unavailable on source domain data \cite{han-eisenstein-2019-unsupervised}.

The majority of domain-adapted models are trained on English domain-specific text, given the availability of English language data. However, many real-world applications, such as working with financial documents \cite{dogu_finbert}, biomedical text \cite{10.1093/bioinformatics/btz682}, and legal opinions and rulings \cite{chalkidis-etal-2020-legal}, should be expected to work in multiple languages. For such applications, annotated target task datasets might be available, but we lack a good pretrained model that we can fine-tune on these datasets.
In this paper, we propose a method for \dapting of a single domain-specific multilingual language model that can be fine-tuned for tasks within that domain in multiple languages. There are several reasons for wanting to train a single model: (i) Data availability: we cannot always find domain-specific text in multiple languages so we should exploit the available resources for effective transfer learning \cite{zhang-etal-2020-improving}. (ii) Compute intensity: it is environmentally unfriendly to domain-adaptive pretrain one model per language \cite{strubell-etal-2019-energy}, and BioBERT was \dapted for 23 days on 8$\times$Nvidia V100 GPUs. (iii) Ease of use: a single multilingual model eases deployment when an organization needs to work with multiple languages on a regular basis \cite{johnson-etal-2017-googles}.

Our method, \mdapting (\MDAPT), extends \dapting to a multilingual scenario, with the goal of training a single multilingual model that performs, as close as possible, to $N$ language-specific models. \MDAPT starts with a base model, i.e. a pretrained multilingual language model, such as mBERT \cite{devlin-etal-2019-bert} or XLM-R \cite{conneau-etal-2020-unsupervised}. As monolingual models have the advantage of language-specificity over multilingual models \cite{rust2020good,ronnqvist-etal-2019-multilingual}, we consider monolingual models as upper baseline to our approach. We assume the availability of English-language domain-specific unlabelled text, and, where possible, multilingual domain-specific text. However, given that multilingual domain-specific text can be a limited resource, we look to Wikipedia for general-domain multilingual text \cite{NEURIPS2019_c04c19c2}. The base model is \dapted on the combination of the domain-specific text, and general-domain multilingual text. Combining these data sources should prevent the base model from forgetting how to represent multiple languages while it adapts to the target domain.

Experiments in the domains of financial text and biomedical text, across seven languages: French, German, Spanish, Romanian, Portuguese, Danish, and English, and on two downstream tasks: named entity recognition, and sentence classification, show the effectiveness of \mdapting.  Further analysis in a cross-lingual biomedical sentence retrieval task indicates that \MDAPT enables models to learn better domain-specific representations, and that these representations transfer across languages. Finally, we show that the difference in tokenizer quality between mono- and multilingual models is more pronounced in domain-specific text, indicating a direction for future improvement.

All models trained with \MDAPT and the new datasets used in downstream tasks and pretraining data\footnote{ \url{https://github.com/RasmusKaer/mDAPT_supplements}} and our code is made available\footnote{ \url{https://github.com/mahartmann/mdapt}}.

\begin{figure}
    \centering
    \includegraphics[width=0.4\textwidth]{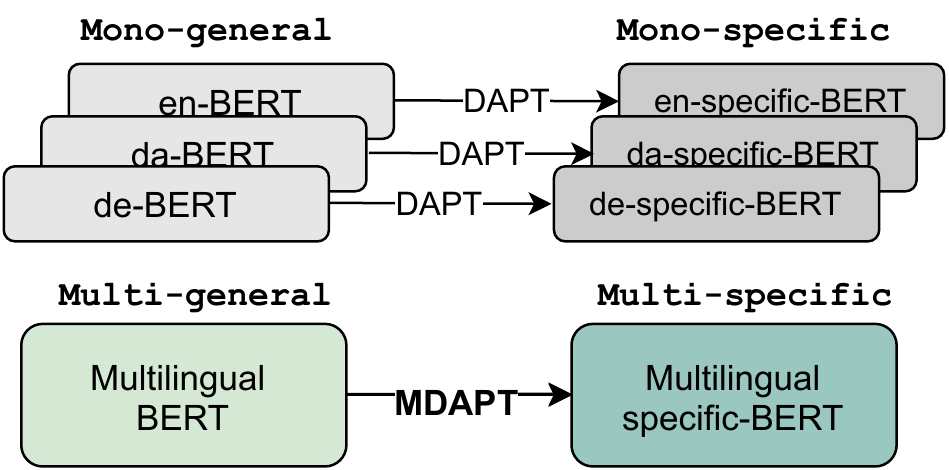}
    \caption{\MDAPT extends \dapting to a multilingual scenario.}
    \label{fig:approach}
\end{figure}



\section{Problem Formulation}

Pretrained language models are trained from random initialization on a large  corpus $\mathcal{C}$ of unlabelled sentences
. Each sentence is used to optimize the parameters of the model using a pretraining objective, for example, masked language modelling, where, for a given sentence, 15\% of the tokens are masked in the input $m$, and the model is trained to predict those tokens $J(\theta) = - \text{log} \; p_{\theta}(x_m \; | \; \mathbf{x}_{\backslash m})$ \cite{devlin-etal-2019-bert}. $\mathcal{C}$ is usually a corpus of no specific domain,\footnote{Text varies along different dimensions, e.g. topic or genre \cite{ramponi-plank-2020-neural}. In the context of this paper, we focus on \emph{domain-specificity} along the topic dimension , i.e. texts are considered as \emph{domain-specific} if they talk about a narrow set of related concepts. The domain-specific text can comprise different genres of text (e.g. financial news articles and financial tweets would both be considered as being from the financial domain).} e.g. Wikipedia or crawled web text.

\textit{Domain-adaptive} pretraining is the process of continuing to pretrain a language model to suit a specific domain \cite{gururangan-etal-2020-dont,han-eisenstein-2019-unsupervised}. This process also uses the masked language modelling pretraining objective, but the model is trained using a domain-specific corpus \specific, e.g. biomedical text if the model should be suited to the biomedical domain. Our goal is to pretrain a \textit{single model}, which will be used for downstream tasks in multiple languages within a specific domain, as opposed to having a separate model for each language. This single multilingual domain-specific model should, ideally, perform as well as language-specific domain-specific models in a domain-specific downstream task.

In pursuit of this goal, we use different types of corpora for \dapting of a single multilingual model. Each considered corpus has two properties: (1) a domain property -- it is a \generalproperty or \specificproperty corpus; and (2) a language property -- it is either \texttt{mono}linugal or \texttt{multi}lingual. These properties can be combined, for example the multilingual Wikipedia is a \multilingual-\generalproperty corpus, while the abstracts of English biomedical publications would be a \monolingual-\specificproperty corpus. Recall that \specificproperty corpora are not always available in languages other than English, but they are useful for adapting to the intended domain; while \multilingual-\generalproperty are more readily available, and should help maintain the multilingual abilities of the adapted language model.
In the remainder of this paper, we will explore the benefits of \dapting with \monolingual-\specificproperty, \multilingual-\specificproperty,  \textit{and} \multilingual-\generalproperty corpora. Figure \ref{fig:approach} shows how \MDAPT extends \dapting to a multilingual scenario.


\section{Multilingual Domain Adaptive Pretraining}\label{sec:approach}
Recall that we assume the availability of large scale English domain-specific and multilingual general unlabelled text. In addition to these  \monolingual-\specificproperty and  \multilingual-\generalproperty corpora, we collect multilingual domain-specific corpora, using two specific domains---financial and biomedical---as an example (Section \ref{section-collecting-domain-specific}).
Note that although we aim to collect domain-specific data in as many languages as possible, the collected data are usually still relatively small.
We thus explore different strategies to combine different data sources (Section~\ref{section-combining-data-sources}), resulting in three different types of pretraining corpora of around 10 million sentences, that exhibit  \specificproperty and \multilingual properties to different extents: \textbf{\adxen}: English domain-specific data; \textbf{\adxmd}: Multilingual domain-specific data, augmented with English domain-specific data; and \textbf{\adxmwiki}: Multilingual domain-specific data, augmented with multilingual general data.

We use mBERT \cite{devlin-etal-2019-bert} as the multilingual base model, and employ two different continued pretraining methods (Section~\ref{section-pretraining-methods}): adapter-based training and full model training, on these three pretraining corpora, respectively.

\begin{table}[t]
    \centering
    \small
    \begin{tabular}{rrcrr}
    \toprule
    \bf Domain & \bf Data & \bf \# Lang. & \bf \# Sent. & \bf \# Tokens \\

    \midrule
    \multirow{3}{*}{Fin} & \textsc{M$_{\text{D}}$} & 14 & \phantom{0}4.9M & \phantom{0}34.4M  \\
    & \textsc{E$_{\text{D}}$} & 1 & 10.0M & 332.8M \\
    & \textsc{M$_{\text{WIKI}}$} & 14 & \phantom{0}5.1M & 199.9M  \\
    
    \midrule
    
    \multirow{3}{*}{Bio} & \textsc{M$_{\text{D}}$} & 8 & \phantom{0}3.2M  & 86.6M\\
    & \textsc{E$_{\text{D}}$} & 1 & 10.0M & 370.6M \\
    & \textsc{M$_{\text{WIKI}}$} & 8 & \phantom{0}6.8M & 214.2M  \\
    \bottomrule
    \end{tabular}
    \caption{A summary of pretraining data used. We use two specific domains---financial (top part) and biomedical (bottom part) as an example in this paper. M stands for Multilingual; E for English; D for Domain-specific; and, Wiki refers to general data, sampled from Wikipedia. The number of tokens are calculated using mBERT cased tokenizer. Note that because languages considered in financial and biomedical domains are not the same , we sample two different \textsc{M$_{\text{WIKI}}$} covering different languages.}
    \label{t:summary_table}
\end{table}

\subsection{Domain-specific corpus}
\label{section-collecting-domain-specific}

\paragraph{Financial domain}
As \specificproperty data for the financial domain, we use
Reuters Corpora (RCV1, RCV2, TRC2),\footnote{Available by request at \url{https://trec.nist.gov/data/reuters/reuters.html}} SEC filings~\cite{desola2019},\footnote{\url{http://people.ischool.berkeley.edu/~khanna/fin10-K}} and \PwCCPT,
which is an in-house collected corpus.
The \PwCCPT consists of articles in multiple languages published on PwC website.
The resulting corpus contains the following languages: \zh, \da, \nl, \fr, \de, \ita, \ja, \no, \pt, \ru, \es, \sv, \en, \tr. 
Statistics on the presented languages can be found in Table~\ref{t:stats:financial} in the Appendix. Information about preprocessing are detailed in Appendix \ref{sec:appendix:preprocfin}.

    
    
    
    

\paragraph{Biomedical domain}
As \specificproperty data for the biomedical domain, we use biomedical publications from the PubMed database,
in the following languages: \fr, \en, \de, \ita, \es, \ro, \ru, \pt.  For languages other than English, we use the language-specific PubMed abstracts published as training data by WMT, and additionally retrieve all language specific paper titles from the database.\footnote{We use data from a bulk download of \url{ftp://ftp.ncbi.nlm.nih.gov/pubmed/baseline}, version 12/14/20} For English, we only sample abstracts. We make sure that no translations of documents are included in the pretraining data. The final statistics on biomedical pretraining data can be found in Table \ref{t:stats:biomedical} in the Appendix, as well as more details about preprocessing the documents. 
The descriptive statistics of these pretraining data can be found in Table~\ref{t:summary_table}.

\subsection{Combination of data sources}
\label{section-combining-data-sources}
\begin{figure}[t]
    \centering
    \includegraphics[width=0.45\textwidth]{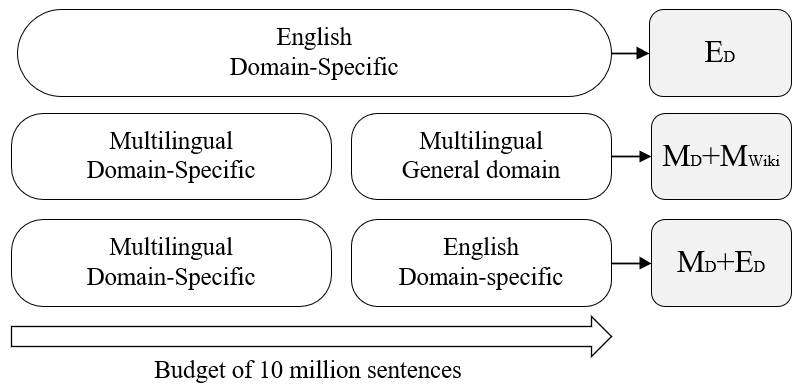}
    \caption{Composition of pretraining data.}
    \label{fig:pretrain_data_combi}
\end{figure}

Recall that \multilingual-\specificproperty data is usually difficult to obtain, and we explore different strategies to account for this lack.
The different compositions of pretraining data are illustrated in Figure \ref{fig:pretrain_data_combi}.
We control the size of the resulting corpora by setting a budget of 10 million sentences. 
This allows a fair comparison across data settings.

With plenty of English \specificproperty text available, \adxen and \adxmd are composed by simply populating the corpus until reaching the allowance. 

As a resource for \multilingual-\generalproperty data, we use Wikipedia page content, where we ensure the same page is not sampled twice across languages. 
Up-sampling \adxmwiki using general domain multilingual data requires a sampling strategy that accounts for individual sizes.
Sampling low-resource languages too often may lead to overfitting the repeated contents, whereas sampling high-resource language too much can lead to a model underfit. 
We balance the language samples using exponentially smoothed weighting~\cite{xue_mt5,NEURIPS2019_c04c19c2,devlin-etal-2019-bert}. 
Following \citeauthor{xue_mt5}, we use a $\alpha$ of 0.3 to smooth the probability of sampling a language, $P(L)$, by $P(L)^\alpha$.
After exponentiating each probability by $\alpha$, we normalize and populate the pretraining corpus with Wikipedia sentences according to smoothed values until reaching our budget.
Except for English, we up-sample using Wikipedia data.  
The statistics of the extracted sentences is presented in tables \ref{t:stats:biomedical} and \ref{t:stats:financial} in the Appendix.



\subsection{Pretraining methods}
\label{section-pretraining-methods}

\paragraph{Continue pretraining the whole model}
We initialize our models with pretrained \base model weights\footnote{\basembert: https://huggingface.co/bert-base-multilingual-cased} and then continue pretraining the whole \base model via the masked language modeling objective. We follow \citet{devlin-etal-2019-bert} in randomly masking out 80\% of subtokens and randomly replacing 10\% of subtokens. For all models, we use an effective batch size of 2048 via gradient accumulation, a sequence length of 128, and a learning rate of 5e-5. We train all models for 25,000 steps, which takes 10 GPU days.

\paragraph{Adapter-based training}
In contrast to fine-tuning all weights of the \base model, adapter-based training introduces a small network between each layer in the \base model, while keeping the \base model fixed.
The resulting adapter weights, which can be optimized using self-supervised pretraining or later downstream supervised objectives, are usually much lighter than the \base model, enabling parameter efficient transfer learning~\citep{houlsby2019parameter}.
We train each adapter for 1.5M steps, taking only 2 GPU days.
We refer readers to~\citet{pfeiffer-etal-2020-mad} for more details of adapter-based training and also describe them in the Appendix~\ref{appendix-adapter-based} for self-containedness.

\section{Domain-Specific Downstream Tasks}
\label{section-evaluation-downstream}
To demonstrate the effectiveness of our multilingual domain-specific models, we conduct experiments on two downstream tasks---Named Entity Recognition (NER) and sentence classification---using datasets from biomedical and financial domains, respectively.

\subsection{NER in the biomedical domain}

\paragraph{Datasets} We evaluate on 5 biomedical NER datasets in different languages. The French \quaero~ \cite{neveol14quaero} dataset, the Romanian \bioro dataset \cite{mitrofan2017bootstrapping}, and the English \ncbi dataset \cite{dougan2014ncbi} comprise biomedical publications. The Spanish \pharmaconer~ \cite{agirre2019pharmaconer} dataset comprises publicly available clinical case studies, and the Portuguese \ptner dataset is the publicly available subset of the data collected by \citet{lopes-etal-2019-contributions}, comprising texts about neurology from a clinical journal. The descriptive statistics of the NER datasets are listed in Table \ref{t:biomedical_ner}, and more details about the datasets can be found in Appendix \ref{sec:appendix:preprocbio}.
We convert all NER annotations to BIO annotation format, and use official train/dev/test splits if available. For \ncbi, we use the data preprocessed by \citet{10.1093/bioinformatics/btz682}. Further preprocessing details can be found in Appendix \ref{sec:appendix:preprocbio}.

\begin{table}[t]
    \centering
    \small
    \setlength{\tabcolsep}{2pt} 
    \begin{tabular}{rrrrrrr}
    \toprule
    & & \bf \ncbishort & \bf \pharmaconershort & \bf \quaero & \bf \ptnershort & \bf \bioro \\
    & & \en & \es & \fr & \pt & \ro \\
    \midrule
\multirow{3}{*}{\# sents.} & train & 5,424 & 8,137 & 1,540 & 1,192 & 1,886 \\
& dev & 923 & 3,801 & 1,481 & 336 & 631 \\
& test & 940 & 3,982 & 1,413 & 973 & 629 \\
\hline
\multirow{3}{*}{\# mentions} & train & 5,134 & 3,810 & 4,516 & 7,600 & 5,180 \\
 & dev & 787 & 1,926 & 4,123 & 2,047 & 1,864 \\
 & test & 960 & 1,876 & 4,086 & 6,315 & 1,768 \\
\hline
\# classes & & 1 & 4 & 10 & 13 & 4 \\
    \bottomrule
    \end{tabular}
    \caption{The descriptive statistics of the biomedical NER datasets.}
    \label{t:biomedical_ner}
\end{table}

\paragraph{NER Model}
Following~\citet{devlin-etal-2019-bert}, we build a linear classifier on top of the BERT encoder outputs, i.e. the contextualized representations of the first sub-token within each token are taken as input to a token-level classifier to predict the token’s tag.
For full model fine-tuning, we train all models for a maximum of 100 epochs, stopping training early if no improvement on the development set is observed within 25 epochs. We optimize using AdamW, a batch size of 32, maximum sequence length of 128,  and a learning rate of 2e-5. 
For adapter-based training, we train for 30 epochs using a learning rate of 1e-4.


\subsection{Sentence classification in the financial domain}
\paragraph{Datasets} We use three financial classification datasets, including the publicly available English \FPB~\cite{Malo2014}, German \GMPC~\cite{Schabus2017}, and a new Danish \DanFin.
The \FPB is an English sentiment analysis dataset where sentences extracted from financial news and company press releases are annotated with three labels (Positive, Negative, and Neutral). Following its annotation guideline, we create \DanFin---a dataset of Danish financial news headlines annotated with a sentiment. 2 annotators were screened to ensure sufficient domain and language background. The resulting dataset has a high inter-rater reliability (a measure of 82.1\% percent agreement for raters and a Krippendorff's alpha of .725, measured on 800 randomly sampled examples).
\GMPC is sourced from an Austrian newspaper. We use \textsc{title} and \textsc{topic} for two classification settings on this dataset: a binary classification, determining whether a \textsc{title} concerns a financial \textsc{topic} or not; and a multi-class classification that classify a \textsc{title} into one of 9 \textsc{topic}s.
We list the descriptive statistics in Table \ref{t:financial_clf}, and further details can be found in Appendix \ref{sec:appendix:ClassFin}.
\begin{table}[t]
    \centering
    \small
    \begin{tabular}{rrrr}
    \toprule
    & \bf \GMPCexp & \bf \DanFin & \bf \FPBtable \\
    & \de & \da & \en \\
    \midrule
    \# sentences & 10,276 & 5,134 & 4,845 \\
    \# classes & 2/9 & 3 & 3 \\
    \bottomrule
    \end{tabular}
    \caption{The descriptive statistics of the financial classification datasets. We frame the German dataset as a binary and a multi-class (9) classification tasks.}
    \label{t:financial_clf}
\end{table}

\paragraph{Classifier}
Following~\citet{devlin-etal-2019-bert}, we built a classification layer on top of the [CLS] token.
We perform simple hyperparameter tuning with the baseline monolingual model on each dataset separately. The parameter setting is selected on a coarse grid of batch-sizes $[16, 32]$ and epochs $[2,4,6]$. The best-performing hyperparameters on each dataset are then used in experiments using other pretrained models.
All experiments follow an 80/20 split for train and testing with an equivalent split for model selection.

\section{Results}\label{sec:results}
\begin{table*}[!ht]
\resizebox{\linewidth}{!}{
    \centering
    \begin{tabular}{ r ccccccccc}
    \toprule
    & \multicolumn{5}{c}{\textsc{Biomedical NER}} &  \multicolumn{4}{c}{\textsc{Financial Sentence Classification}} \\
    \cmidrule{2-10} 
    & \bf \quaero & \bf \bioro & \bf \pharmaconershort & \bf \ncbishort & \bf \ptner  & \bf \omptwotable & \bf \ompninetable & \bf \DanFintable & \bf \FPBtable \\ 
    & \fr & \ro & \es & \en & \pt & \de & \de & \da & \en \\ 
    \cmidrule{1-10} 
    & \multicolumn{9}{c}{\textsc{Full model pretraining}} \\
      \cmidrule{1-10} 
    \monolingual-\specificproperty-\bert & - & - & - & 88.1& 72.9&-&-&-& 87.3\\
    \cmidrule{1-10} 
\monolingual-\bert & \bf 61.9& \bf 75.5 &88.2 &85.1& 72.6 & \bf 91.4 & 71.5 & \bf 65.2 & \bf 85.0 \\   
    
    
   \basembert & -3.7& -1.6&+0.2 &+1.0 & -0.2 & -0.6 & -0.4 & -2.4 & -2.6 \\ 
   
    
 + \adxen & -3.6 &-1.6 &\bf+0.6 & +1.5& -0.6& -0.3 & 0 & -2.5 & -1.2 \\

    
     + \adxmd & -2.7&-0.9 & +0.5 & \bf +2.1& \bf +0.1& -0.2 & \bf +0.1 & -1.6 & -1.1 \\

    
     + \adxmwiki & -2.1&-1.4  & +0.3  & +1.8&0.0& -0.1 & \bf +0.1 & -1.6 & -1.4 \\

      \cmidrule{1-6} \cmidrule{7-10}
& \multicolumn{9}{c}{\textsc{Adapter-based pretraining}} \\
  \cmidrule{2-10} 
\monolingual-\bert & \bf 58.6 & \bf 73.2 & 86.6 & 82.6 & 63.5 & 90.5 & 69.1 & \bf 66.0 & \bf 85.3 \\
\basembert & -4.5 & -4.5 & -0.3 & +0.1 & -3.7 & 0.0 & +0.8 & -3.1 & -3.1 \\  
 + \adxen & -2.9 & -2.0 & +1.5 & +1.4 & +1.8 & +0.7 & +1.5 & -4.9 & -3.5 \\  
 + \adxmd & -1.3 & -1.9 & \bf +1.9 & +1.4 & \bf +2.7 & \bf +0.9 & \bf +3.8 & -1.7 & -2.6 \\ 
 + \adxmwiki & -1.4 & -2.6 & +1.0 & \bf +1.8 & +1.6 & +0.6 & +2.6 & -1.9 & -3.2 \\ 
\bottomrule
    \end{tabular}}
    \caption{Evaluation results on biomedical NER and financial sentence classification tasks. We report the results---span-level micro $F_1$ for NER and sentence-level micro $F_1$ for classification---on the monolingual BERTs. Performance differences compared to the monolingual baselines are reported for multilingual BERTs, with and without \MDAPT. All experiments are repeated five times using different random seeds, and mean values are reported.}
    \label{table-main-results}
\end{table*}

To measure the effectiveness of \mdapting, we compare the effectiveness of our models trained with \MDAPT on downstream NER and classification, to the respective monolingual baselines (\monolingual-\generalproperty), and to the base multilingual model without \MDAPT (Table~\ref{table-main-results}). Where available, we also compare to the respective monolingual domain-specific models (\monolingual-\specificproperty).
\paragraph{Baseline models}
As \monolingual-\generalproperty baselines, we use English BERT~\cite{devlin-etal-2019-bert}, Portuguese BERT~\cite{souza2020bertimbau}, Romanian BERT~\cite{dumitrescu-etal-2020-birth}, BETO~\cite{CaneteCFP2020} for Spanish, FlauBert  \cite{le-etal-2020-flaubert} for French, German BERT~\cite{chan-etal-2020-germans}, and Danish BERT.\footnote{\url{https://github.com/botxo/nordic_bert}} 
\monolingualupper-\specificproperty baselines exist only for a few languages and domains, we use \biobert \cite{10.1093/bioinformatics/btz682} as English biomedical baseline, and \finbert \cite{dogu_finbert} as English financial baseline. To the best of our knowledge, \ptbiobert \cite{schneider-etal-2020-biobertpt} is the only biomedical model for non-English language, we use it as Portuguese biomedical baseline, see Appendix \ref{sec:appendix:baselines} for more details. 


\subsection{Main results}
The main results for the biomedical NER and financial sentence classification tasks are presented in Table \ref{table-main-results}. We report the evaluation results for the \monolingual-\bert baselines in the respective languages and the performance difference of the multilingual models compared to these monolingual baselines. 
We also consider two domain adaptive pretraining approaches: full model training, reported in the upper half of the table, and adapter-based training in the lower half. 

Our work is motivated by the finding that \dapting enables models to better solve domain-specific tasks in monolingual scenarios. The first row in Table \ref{table-main-results} shows our re-evaluation of the performance of the three available \dapted \monolingual-\specificproperty-\bert models matching the domains investigated in our study. We confirm the findings of the original works, that the domain-specific models outperform their general domain \monolingual-\bert counterparts. This underlines the importance of domain adaptation in order to best solve domain-specific task. The improvements of \ptbiobert over \bertpt are small, which coincides with the findings of \citet{schneider-etal-2020-biobertpt}, and might be due to the fact that the \ptner dataset comprises clinical entities rather than more general biomedical entities.

\paragraph{Full model training}
Recall that the aim of \MDAPT is to train a single \multilingual-\specificproperty model that performs comparable to the respective \monolingual-\generalproperty model. Using full model pretraining, we observe that the \dapted multilingual models can even outperform the monolingual baselines for \es and \en biomedical NER, and \de for financial sentence classification. On the other hand, we observe losses of the multilingual models over the monolingual baselines for \fr and \ro NER, and \da and \en sentence classification. In all cases, \MDAPT narrows the gap to monolingual performance compared to \basembert, i.e. \mdapting helps to make the multilingual model better suited for the specific domain.

\paragraph{Adapter-based training}
Adapter-based training exhibits a similar pattern: \MDAPT improves \basembert across the board, except for the \da and \en sentence classification tasks, where \MDAPT is conducted using only \en-\specificproperty data. 
For most tasks, except \da and \en sentence classification, the performance of adapter-based training is below the one of full model training.
On \pt NER dataset, the best score (66.2) achieved by adapter-based training is much lower than the one (72.7) by the full model training.

\begin{table}[t]
\resizebox{\linewidth}{!}{
    \centering
    \begin{tabular}{rccc}
    \toprule
    &\basembert & \MDAPT & $\neg$ \MDAPT\\
    \midrule
    \quaero & 58.2& 59.8& 58.0\\
    \bioro & 73.9 & 74.5 & 73.4\\
    \ncbishort & 86.0& 87.2& 85.9\\
    \ptnershort &72.4& 72.7&71.8\\
    \pharmaconershort & 88.5 & 88.9 & 87.8\\
    \FPBtable & 82.4 & 83.9 & 82.5 \\
    \DanFintable & 62.8 & 63.6 & 62.2 \\
    \omptwotable & 90.8 & 91.3 & 91.0 \\
    \ompninetable & 71.1 & 71.6 & 71.0/71.7\\
    \bottomrule
    \end{tabular}}
    \caption{Cross-domain control experiments. We report two control results for \ompninetable since two \MDAPT-setting achieved the same averaged accuracy.}
    \label{t:control_experiments}
\end{table}



\paragraph{Comparison of combination strategies}
After we observe a single \multilingual model can achieve competitive performance as several \monolingual models, the next question is how do different combination strategies affect the effectiveness of \MDAPT?
As a general trend, the pretraining corpus composed of multilingual data---\adxmd and \adxmwiki---achieves better results than \adxen composed by only \en data. This is evident across both full - and adapter-based training.  
\adxmd performs best in most cases, especially for the adapter-based training. 
This result indicates the importance of multilingual data in the pretraining corpus.
It is worth noting that even pretraining only on \adxen data can improve the performance on non-English datasets, and for \en tasks, we see an expected advantage of having more \en-\specificproperty data in the corpus. 

\subsection{Cross-domain evaluations}
To make sure that the improvements of \MDAPT models over \basembert stem from observing multilingual domain-specific data, and not from exposure to more data in general, we run cross-domain experiments \cite{gururangan-etal-2020-dont}, where we evaluate the models adapted to the biomedical domain on the financial downstream tasks, and vice versa. The results are shown in Table \ref{t:control_experiments}, where we report results for the best \MDAPT model and its counterpart in the other domain ($\neg$ \MDAPT). In almost all cases, \MDAPT outperforms $\neg$ \MDAPT, indicating that adaptation to the domain, and not the exposure to additional multilingual data is responsible for \MDAPT's improvement over \basembert. For the \textsc{omp} datasets, $\neg$ \MDAPT performs surprisingly well, and we speculate this might be because it requires less domain-specific language understanding to classify the newspaper titles.

\section{Analysis}\label{sec:analysis}
Our experiments suggest that \MDAPT results in a pretrained model which is better suited to solve domain-specific downstream tasks than \basembert, and that \MDAPT narrows the gap to monolingual model performance. In this section, we present further analysis of these findings, in particular we investigate the quality of domain-specific representations learned by \MDAPT models compared to \basembert, and the gap between mono- and multilingual model performance.

\paragraph{Domain-specific multilingual representations}
Multilingual domain adaptive pretraining should result in improved representations of domain-specific text in multiple languages. We evaluate the models' ability to learn better sentence representations via a cross-lingual sentence retrieval task, where, given a sentence in a source language, the model is tasked to retrieve the corresponding translation in the target language. To obtain a sentence representation, we average over the encoder outputs for all subtokens in the sentence, and retrieve the k nearest neighbors based on cosine similarity. As no fine-tuning is needed to perform this task, it allows to directly evaluate encoder quality. We perform sentence retrieval on the parallel test sets of the WMT Biomedical Translation Shared Task 2020 \cite{bawden-etal-2020-findings}. The results in Table \ref{t:sentence_retrieval} show that \MDAPT improves retrieval quality, presumably because the models learned better domain-specific representations across languages. Interestingly, with English as target language (upper half), the model trained on English domain-specific data works best, whereas for English as source language, it is important that the model has seen multilingual domain-specific data during pretraining.

\begin{table}[t]
\resizebox{\linewidth}{!}{
    \centering
    \begin{tabular}{lcccc}
\toprule
&\basembert& +\adxen & + \adxmd &+ \adxmwikishort \\
\midrule
\es $\to$ \en & 86.7  & \textbf{91.9}  & 89.4  & 87.2 \\
\pt $\to$ \en  & \textbf{87.3} & 77.1  & 77.5 & 83.9 \\
\de $\to$ \en  & 79.4  & \textbf{88.7}  &   83.9& 80.9  \\
\ita $\to$ \en  & 85.6  & \textbf{90.9}  & 87.4  & 87.1  \\
\ru $\to$ \en  & 67.5  & \textbf{84.4}  & 76.5  & 74.6 \\
\midrule
\en $\to$ \es & 86.7  & 84.7 & \textbf{90.5}  & 87.4 \\
\en $\to$ \pt & 89.4 & 78.2 &    \textbf{90.4}&  86.8\\
\en $\to$ \de &  79.4&   79.6&   \textbf{87.8}&   81.2\\
\en $\to$ \ita &  83.9&   82.9&    \textbf{88.1}& 86.1 \\
\en $\to$ \ru &  70.3&  81.6 &    \textbf{90.8}&  89.5\\
\bottomrule
    \end{tabular}}
    \caption{Precision@1 for biomedical sentence retrieval. Best score in each row is marked in bold. The upper half shows alignment to English, the lower half alignment from English.}
    \label{t:sentence_retrieval}
\end{table}

\begin{figure}[h!]
    \centering
    \includegraphics[width=0.4\textwidth]{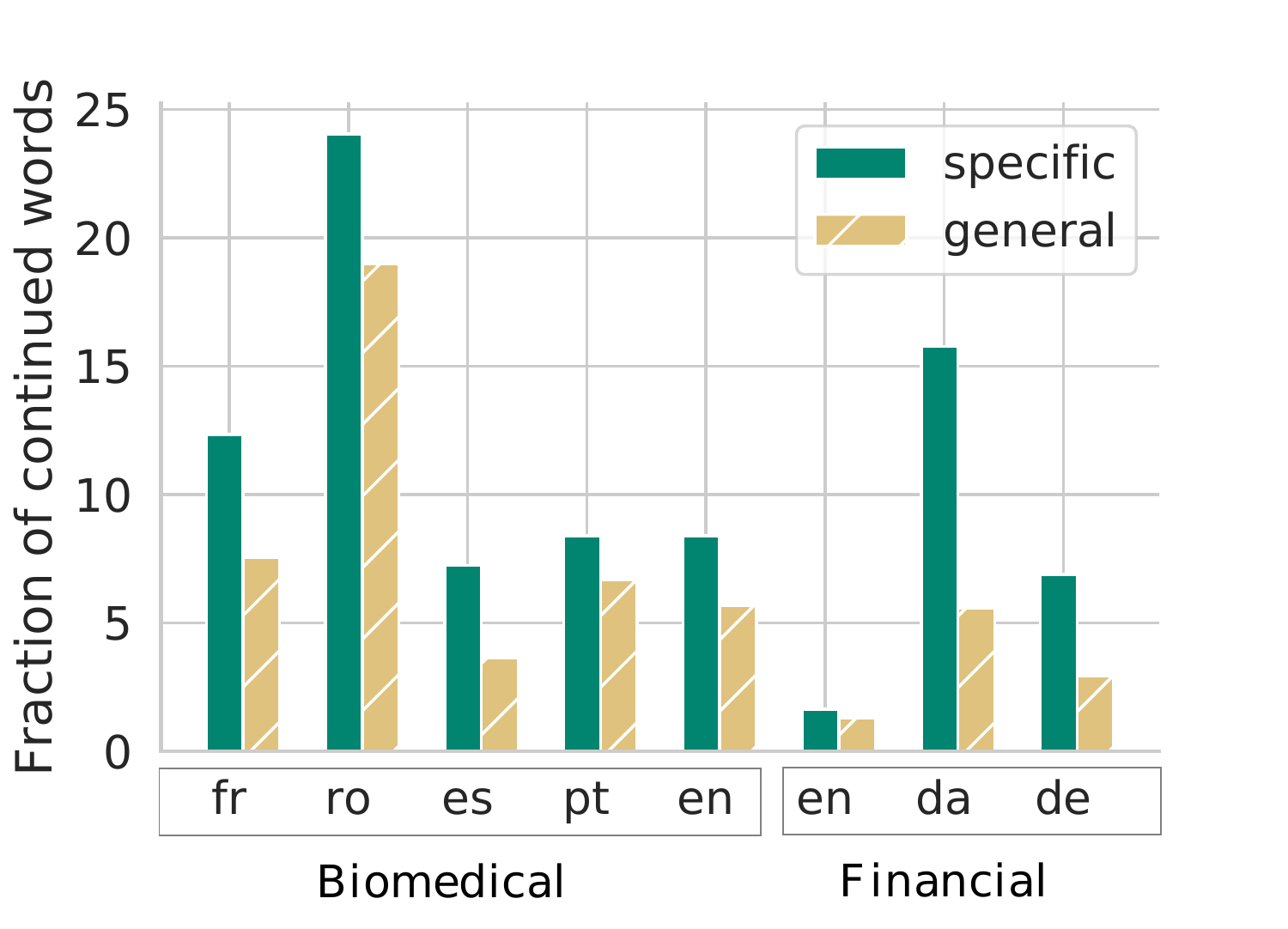}
    \caption{Difference in fraction of continued words between \monolingual - and \multilingual-lingual tokenizers on general and specific datasets. The bars indicate improvement of the monolingual tokenizer over the multilingual tokenizer. }
    \label{fig:tokenization}
\end{figure}

 \paragraph{Effect of tokenization} 
Ideally, we want to have a \multispecificmodel model that performs close to the corresponding monolingual model. However, for the full fine-tuning setup, the monolingual model outperforms the \multispecificmodel models in most cases. \citet{rust2020good} find that the superiority of monolingual over multilingual models can partly be attributed to better tokenizers of the monolingual models, and we hypothesize that this difference in tokenization is even more pronounced in domain-specific text.
Following \citet{rust2020good}, we measure tokenizer quality via \emph{continued words}, the fraction of words that the tokenizer splits into several subtokens, and compare the difference between monolingual and multilingual tokenizer quality on \specificproperty text (the train splits of the downstream tasks), with their difference on \generalproperty text sampled from Wikipedia. Figure \ref{fig:tokenization} shows that the gap between monolingual and multilingual tokenization quality is indeed larger in the \specificproperty texts (green bars) compared to the \generalproperty texts (brown bars), indicating that in a specific domain, it is even harder for a multilingual model to outperform a monolingual model. This suggests that methods for explicitly adding representations of domain-specific words \cite{poerner-etal-2020-inexpensive,schick-schutze-2020-bertram} could be a promising direction for improving our approach. 
 
\paragraph{Error analysis on financial sentence classification}
To provide a better insight into the difference between the \monolingual and \multilingual models, we compare the error predictions on the Danish \DanFin dataset, since results in Table~\ref{table-main-results} show that the \monolingual outperforms all \multilingual models with a large margin on this dataset.
We note that the \DanFin dataset, which is sampled from tweets, contains a heavy use of idioms and jargon, on which the \multilingual models usually fail. 
For example,
\begin{itemize}
    \par
    \item Markedet lukker: \textbf{Medvind} til bankaktier på en rød C25-dag [\textsc{Positive}]
    \par
    English translation: \textit{Market closes: \textbf {Tailwind} for bank shares on a red C25-day}
    \item Nationalbanken tror ikke særskat får den store betydning: Ekspert kaldet det \textbf{"noget pladder"} [\textsc{Negative}]
    \par
    English translation: \textit{The Nationalbank does not think special tax will have the great significance: Expert called it \textbf{"some hogwash"}}
\end{itemize}

Pretraining data for the \monolingual \bertda includes Common Crawl texts and custom scraped data from two large debate forums. We believe this exposes the \bertda to the particular use of informal register.
By contrast, the pretraining data we use are mainly sampled from publications. 
This could be an interesting direction of covering the variety of a language in sub-domains for a strong \multispecificmodel model. 


\section{Related Work}
Recent studies on domain-specific \bert~\citep{10.1093/bioinformatics/btz682,alsentzer-etal-2019-publicly,nguyen-etal-2020-bertweet}, which mainly focus on English text, have demonstrated that in-domain pretraining data can improve the effectiveness of pretrained models on downstream tasks. 
These works continue pretraining the whole \base model---\bert or \roberta---on domain-specific corpora, and the resulting models are supposed to capture both generic and domain-specific knowledge.
By contrast, \citet{beltagy-etal-2019-scibert,gu2020domain,shin-etal-2020-biomegatron} train domain-specific models from scratch, tying an in-domain vocabulary. 
Despite its effectiveness, this approach requires much more compute than \dapting, which our work focuses on. 
Additionally, we explore an efficient variant of \dapting based on adapters~\citep{houlsby2019parameter,pfeiffer-etal-2020-mad}, and observe similar patterns regarding pretraining a multilingual domain-specific model.


Several efforts have trained large scale multilingual language representation models using parallel data~\citep{aharoni-etal-2019-massively,NEURIPS2019_c04c19c2} or without any cross-lingual supervision~\citep{devlin-etal-2019-bert,conneau-etal-2020-unsupervised,xue_mt5}.
However, poor performance on low-resource languages is often observed, and efforts are made to mitigate this problem~\citep{rahimi-etal-2019-massively,ponti2020parameter,pfeiffer-etal-2020-mad}.
In contrast, we focus on the scenario that the NLP model needs to process domain-specific text supporting a modest number of languages. 

Alternative approaches aim at adapting a model to a specific target task within the domain directly, e.g. by an intermediate supervised fine-tuning step \cite{pruksachatkun-etal-2020-intermediate, phang-etal-2020-english}, resulting in a model specialized for a single task. Domain adaptive pretraining, on the other hand, aims at providing a good base model for different tasks within the specific domain. 

\section{Conclusion}
We extend domain adaptive pretraining to a multilingual scenario that aims to train a single multilingual model better suited for the specific domain.
Evaluation results on datasets from biomedical and financial domains show that although multilingual models usually underperform their monolingual counterparts, domain adaptive pretraining can effectively narrow this gap. 
On seven out of nine datasets for document classification and NER, the model resulting from multilingual domain adaptive pretraining outperforms the baseline \multilingual-\generalproperty model, and on four it even outperforms the \monolingual-\generalproperty model.
The encouraging results show the implication of deploying a single model which can process financial or biomedical documents in different languages, rather than building separate models for each individual language. 
\section*{Acknowledgements}
We thank PwC for providing the data and thank the experts for their annotation support and valuable feedback. Mareike Hartmann was funded by the Lundbeck foundation for research in the context of the BrainDrugs project and the German Federal Ministry of Research (BMBF) in the context of the XAINES project.
Xiang Dai and Desmond Elliott are supported by Innovation Fund Denmark in the context of AI4Xray project.

\bibliography{anthology,custom}
\bibliographystyle{acl_natbib}

\clearpage
\newpage
\clearpage

\appendix
\section{Baseline models}\label{sec:appendix:baselines}
Table~\ref{table-list-bert-models} is a comparison between baseline \monolingual models and the \multilingual model. For the NER tasks, we use the \texttt{cased} versions for all experiments. For sentence classification, we use \texttt{uncased} versions for \bertda and \berten.

\begin{table*}[t]
    \centering
    \begin{tabular}{p{4cm}p{5.5cm}cc}
    \toprule
    & Training data & Vocab size & \# parameters \\
    \midrule
\bertde~\cite{chan-etal-2020-germans} & OSCAR (Common Crawl), OPUS (Translated web texts), Wikipedia, Court decisions [163.4G] & 30.0K & 109.1M \\ 
\hline
\bertda & Common Crawl, Wikipedia, Debate forums, OpenSubtitles [9.5G, 1.6B] & 31.7K & 110.6M \\
\hline
\berten~\cite{devlin-etal-2019-bert} & English Wikipedia, Books [3.3B] & 29.0K & 108.3M \\
\hline
\biobert~\citep{10.1093/bioinformatics/btz682} & Initialized with \berten; continue on PubMed, PMC [18B] & \multirow{2}{*}{29.0K} & \multirow{2}{*}{108.3M} \\
\hline
\finbert~\cite{dogu_finbert} & Initialized with \berten; continue on News articles [29M] & 30.5K & 109.5M \\
\hline
\bertes~\cite{CaneteCFP2020} & OPUS, Wikipedia [3B] & 31.0K & 109.9M \\
\hline
\bertfr~\cite{le-etal-2020-flaubert} & 24 corpora, including CommonCrawl, Wikipedia, OPUS, Books, News, and data from machine translation shared tasks, Wikimedia projects [71G, 12.7B] & 68.7K & 138.2M \\
\hline
\bertpt~\cite{souza2020bertimbau} & brWaC (web text for Brazilian Portuguese) [2.6B] & 29.8K & 108.9M \\
\hline
\ptbiobert~\citep{schneider-etal-2020-biobertpt} & Initialized with \basembert; continue on PubMed and Scielo (scholarly articles) [16.4M] & 119.5K & 177.9M \\
\hline
\bertro~\cite{dumitrescu-etal-2020-birth} & OSCAR, OPUS, Wikipedia [15.2G, 2.4B] & 50.0K & 124.4M \\
\midrule
\basembert~\cite{devlin-etal-2019-bert} & Wikipedia [72G] & 119.5K & 177.9M \\
    \bottomrule
    \end{tabular}
    \caption{A comparison between baseline \monolingual models and the \multilingual model: \basembert. We use total file size (Gigabyte) and the total number of tokens to represent the training data size.}
    \label{table-list-bert-models}
\end{table*}enclose

\section{Biomedical data}\label{sec:appendix:preprocbio}
\paragraph{Preprocessing pretraining data}
For the English abstracts, we sentence tokenize using NLTK and filter out sentences that do not contain letters. For the WMT abstracts, we filter out lines that start with \#, as these indicate paper ID and author list. We determine the language of a document using its metadata provided by PubMed. We transliterate Russian PubMed titles (in Latin) back to Cyrillic using the \texttt{transliterate} python package (\url{https://pypi.org/project/transliterate/}).

\paragraph{Downstream NER data}
The French \quaero~ \cite{neveol14quaero} dataset comprises titles of research articles indexed in the biomedical MEDLINE database, and information on marketed drugs from the European Medicines Agency. The Romanian \bioro~ \cite{mitrofan2017bootstrapping} dataset consists of biomedical publications across various medical disciplines. The Spanish \pharmaconer~ \cite{agirre2019pharmaconer} dataset comprises publicly available clinical case studies, which show properties of the biomedical literature as well as clinical records, and has annotations for pharmacological substances, compounds and proteins. The English \ncbi~ \cite{dougan2014ncbi} dataset consists of PubMed abstracts annotated for disease names. The Portuguese \ptner dataset is the publicly available subset of the data collected by \citet{lopes-etal-2019-contributions}, and comprises texts about neurology from a clinical journal.

\subparagraph{Prepocessing NER data}
We convert all annotations to BIO format. The gaps in discontinuous entities are labeled. We sentence tokenize at line breaks, and if unavailable at fullstops. We word tokenize all data at white spaces and split off numbers and special characters. If available, we use official train/dev/test splits. For \bioro, we produce a random 60/20/20 split. For \ptner, we use the data from volume 2 for training and development data and test on volume 1.





\begin{table*}[t]
\centering
\begin{tabular}{rrrrr}
\toprule
Lang & PM abstracts & PM titles & \textsc{M$_{\text{D}}$} & \textsc{M$_{\text{WIKI}}$} \\
\midrule
fr&	54,047&	681,774&	735,821 & 872,678\\
es&	73,704&	312,169&	385,873 & 939,452\\
de&	31,849&	814,158&	846,007&831,257\\
it&	14,031&	265,272&	279,303&923,548\\
pt&	38,716&	79,766&	118,482& 811,522\\
ru&	43,050&	576,684&	619,734&908,011 \\
ro&	0&	27,006&	27,006&569,792\\
en&		227,808	&0& 227,808&903,706\\
\midrule
Total &	483,205&	2,756,829&	3,240,034&6,759,966\\
\bottomrule
 \end{tabular}
 \caption{Number of sentences of multilingual domain-specific pre-training data for biomedical domain. Upsampling for \en was done from PM abstracts instead of Wikipedia.}\label{t:stats:biomedical}
 \end{table*}

\begin{table*}[t]
\centering
\begin{tabular}{rrrrr}
\toprule
Lang &	RCV2 &	PwC & \textsc{M$_{\text{D}}$} & \textsc{M$_{\text{WIKI}}$} \\
\midrule
	zh &	222,308 &	1,466 &	223,774 &	470,111 \\
	da &	72,349 &	192,352 &	264,701 &	465,044 \\
	nl &	15,131 &	34,344 &	49,475 &	391,750 \\
	fr &	863,911 &	51,500 &	915,411 &	143,427 \\
	de &	1,104,603 &	71,382 &	1,175,985 &	0 \\
	it &	138,814 &	22,499 &	161,313 &	467,680 \\
	ja &	88,333 &	20,936 &	109,269 &	450,352 \\
	no &	92,828 &	19,208 &	112,036 &	451,799 \\
	pt &	57,321 &	35,323 &	92,644 &	439,942 \\
	ru &	192,869 &	48,388 &	241,257 &	468,466 \\
	es &	936,402 &	51,100 &	987,502 &	95,691 \\
	sv &	132,456 &	25,336 &	157,792 &	467,050 \\
	en &	0 &	346,856 &	346,856 &	444,532 \\
	tr &	0 &	34,990 &	34,990 &	362,685 \\
\midrule
Total &	3,917,325 &	955,680 & 4,873,005 &	5,118,529 \\
\bottomrule
 \end{tabular}
 \caption{Number of sentences of multilingual domain-specific pretraining data for financial domain. Upsampling for \en used the TRC2 corpus instead of Wikipedia.}\label{t:stats:financial}
 \end{table*}

\section{Financial data}\label{sec:appendix:ClassFin}
\paragraph{Preprocessing pretraining data}\label{sec:appendix:preprocfin}
Sentences are tokenized using NLTK. For languages not cover by the sentence tokenizer, we split by full stops.  
Additionally, a split check of particular large sentences, filtering out sentences with no letters, and HTML and tags have been removed.

    
    

\paragraph{Downstream classification data}
\subparagraph{\PwCCPT}
    The corpus consists of PwC publications in multiple languages made publicly available on PwC websites.
    The publications cover a diverse range of topics that relates to the financial domain.
    The corpus is created by extracting text passages from publications.
    Table \ref{fig:pretrain_data_combi} describes the number of sentences and the languages that the CPT corpus cover.
    
\subparagraph{\DanFin}
    The financial sentiment dataset is curated from financial newspapers headline tweets. 
    The motivation was to create a Danish equivalent to \FPB.
    The news headlines are annotated with a sentiment by 2 annotators.
    The annotators were screened to ensure sufficient domain and educational background. 
    A description of \emph{positive}, \emph{neutral}, and \emph{negative} was formalized before the annotation process.
    The dataset has an 82.125\% rater agreement and a Krippendorff's alpha of .725 measured on 800 randomly sampled instances.

\subparagraph{\GMPC~\cite{Schabus2017}} The annotated dataset includes user comments posted to an Austrian newspaper. We use the \textsc{title} (newspaper headline) and \textsc{Topics}, i.e., \textsc{'Kultur', 'Sport', 'Wirtschaft', 'International', 'Inland', 'Wissenschaft', 'Panorama', 'Etat', 'Web'}. 
With the dataset, we derive two downstream tasks.
The binary classification task \GMPCexp$_{binary}$ that deals with whether a \textsc{title} concerns a financial \textsc{Topics} or not. Here we merge all non-financial \textsc{Topics} into one category.
The multi-class classification \GMPCexp$_{multi}$ seeks to classify a \textsc{title} into one of the 9 \textsc{Topics}. 

\section{Adapter-based training}
\label{appendix-adapter-based}
Recall that the main component of a transformer model is a stack of transformer layers, each of which consists of a multi-head self-attention network and a feed-forward network, followed by layer normalization. The idea of adapter-based training~\citep{houlsby2019parameter,Stickland:Murray:ICML:2019,Pfeiffer:Ruckle:EMNLP:2020} is to add a small size network (called \emph{adapter}) into each transformer layer. Then during the training stage, only the weights of new adapters are updated while keeping the base transformer model fixed. Different options regarding where adapters are placed, and its network architecture exist. In this work, we use the bottleneck architecture proposed by~\citet{houlsby2019parameter} and put the adapters after the feed-forward network, following~\citep{Pfeiffer:Ruckle:EMNLP:2020}:
$$
    \text{Adapter}_l \left(\textit{h}_l, \textit{r}_l \right) = \textit{U}_l \left( \text{ReLU} \left( \textit{D}_l \left( \textit{h}_l \right) \right) \right) + \textit{r}_l
$$
where $\textit{r}_l$ is the output of the transformer's feed-forward layer and $\textit{h}_l$ is the output of the subsequent layer normalisation.



\end{document}